\documentclass[preprint,12pt]{elsarticle}

\usepackage{CJK}
\usepackage[top=2cm, bottom=2cm, left=2cm, right=2cm]{geometry}
\usepackage{algorithm}
\usepackage{algorithmicx}
\usepackage{algpseudocode}
\usepackage{amsmath,amssymb}
\usepackage{lineno,hyperref}
\usepackage{subfigure}
\usepackage{epstopdf}
\usepackage{multicol}
\usepackage{multirow}
\usepackage{comment}
\modulolinenumbers[5]

\usepackage{threeparttable}
\usepackage{enumerate}
\usepackage{booktabs}
\usepackage{epstopdf}
\usepackage{wrapfig}

\usepackage{amsmath}
\usepackage{bm}
\usepackage{booktabs}
\usepackage{graphicx}
\usepackage{resizegather}
\usepackage{pifont}
\usepackage{xcolor} 
\newcommand{\cmark}{\ding{51}}

\journal{Neurocomputing}

\begin{document}

\begin{frontmatter}



\title{\textcolor{black}{HyCTAS: Multi-Objective Hybrid Convolution-Transformer Architecture Search for Real-Time Image Segmentation}}


\author{Hongyuan Yu\(^{1}\), Cheng Wan\(^{2}\), Xiyang Dai\(^{5*}\), Mengchen Liu\(^{5}\), Dongdong Chen\(^{5}\), Bin Xiao\(^{5}\), Yan Huang\(^{3,4}\),  Yuan Lu\(^{5}\),  Liang Wang\(^{3,4}\)}
\address{
\(^{1}\)The Multimedia Department, Xiaomi Inc. (XIAOMI)\\
\(^{2}\)Georgia Institute of Technology (GATECH) \\
\(^{3}\)Center for Research on Intelligent Perception and Computing (CRIPAC),\\
National Laboratory of Pattern Recognition (NLPR) \\
\(^{4}\)University of Chinese Academy of Sciences (UCAS) \\
\(^{5}\)Microsoft Inc. \\
yuhyuan1995@gmail.com\\
cwan38@gatech.edu \\
cddlyf@gmail.com \\
\{yhuang, wangliang\}@nlpr.ia.ac.cn \\
\{mengcliu, dochen, bin.xiao, luyuan, xiyang.dai\}@microsoft.com \\
}
\cortext[mycorrespondingauthor]{Corresponding author}

\begin{abstract}
\textcolor{black}{Real-time image segmentation demands architectures that preserve fine spatial detail while capturing global context under tight latency and memory budgets.} Image segmentation is one of the most fundamental problems in computer vision and has drawn a lot of attention due to its vast applications in image understanding and autonomous driving. However, designing effective and efficient segmentation neural architectures is a labor-intensive process that may require numerous trials by human experts. In this paper, we address the challenge of integrating multi-head self-attention into high-resolution representation CNNs efficiently by leveraging architecture search. Manually replacing convolution layers with multi-head self-attention is non-trivial due to the costly overhead in memory to maintain high resolution. By contrast, we develop a multi-target multi-branch supernet method, which not only fully utilizes the advantages of high-resolution features but also finds the proper location for placing the multi-head self-attention module. Our search algorithm is optimized towards multiple objectives (e.g., latency and mIoU) and is capable of finding architectures on the \textcolor{black}{approximate Pareto front} with an arbitrary number of branches in a single search. We further present a series of models via the Hybrid Convolutional-Transformer Architecture Search (HyCTAS) method that searches for the best hybrid combination of \textcolor{black}{lightweight} convolution layers and memory-efficient self-attention layers between branches from different resolutions and \textcolor{black}{fuses them at high resolution} for both efficiency and effectiveness. \textcolor{black}{On Cityscapes, ADE20K, and COCO, HyCTAS discovers competitive real-time models without ImageNet pretraining, delivering strong accuracy and latency trade-offs. Code and models are available at \url{https://github.com/MarvinYu1995/HyCTAS}}.
\end{abstract}


\end{frontmatter}


\section{Introduction}
Image segmentation predicts pixel-level annotations of different semantic categories for an image. Unlike image classification, which assigns a single label for an entire image, image segmentation requires a per-pixel label, making it a core challenge in tasks such as autonomous driving, medical image analysis, and general scene understanding. Specifically, semantic segmentation assigns each pixel to one of several predefined categories (e.g., sky, tree, human, car), instance segmentation differentiates individual object instances of the same category, and panoptic segmentation~\cite{he2017mask} unifies these two perspectives by assigning each pixel a class label and an instance identity.

Over the past decade, deep neural networks have revolutionized segmentation performance~\cite{zhang2021cross, Cheng2020HigherHRNetSR}. In particular, high-resolution representation architectures (e.g.\ HRNet~\cite{WangSCJDZLMTWLX19}) have significantly boosted accuracy by preserving spatial details throughout the network. However, these improvements often come at the cost of high computational overhead. Manually designing a segmentation framework that retains fine spatial features, incorporates global context (e.g.\ via attention mechanisms), and still runs in real time is notoriously difficult. Recent surveys underline this tension between accuracy and efficiency in modern segmentation systems~\cite{baymurzina2022review,mo2022review}.

\textcolor{black}{Despite the prevalence of hybrid CNN–Transformer designs, many hybrid NAS frameworks restrict the search to single-path or single-resolution spaces and yield a single model per run, which makes it difficult to retain high-resolution streams and to cover diverse latency budgets. HyCTAS addresses this by enabling a multi-branch, multi-resolution search that returns an \emph{approximate Pareto front } of architectures in a single run, explicitly balancing mIoU and latency.}

\textcolor{black}{In this paper, multi-branch refers to parallel feature streams at different spatial resolutions that are maintained throughout the network (in the spirit of HRNet), and multi-objective refers to simultaneous optimization of multiple objectives (e.g., mIoU versus latency/FLOPs/params) under a multi-objective formulation. We use approximate Pareto front/non-dominated set to denote the set of architectures not dominated by any other evaluated candidates in our search space~\cite{deb2002fast}.}

\textcolor{black}{The remainder of the paper is organized as follows: Section~\ref{sec:related} reviews recent segmentation and NAS works. Section~\ref{sec:method} introduces the HyCTAS search space and algorithm. Section~\ref{sec:exp} reports experiments across Cityscapes, ADE20K, and COCO with ablations, followed by limitations and conclusions.}


Recently, transformer-based models~\cite{dosovitskiy2020, SETR} have shown promise in vision tasks by modeling long-range dependencies. Yet, straightforwardly replacing convolutional layers with self-attention can be impractical for high-resolution segmentation due to quadratic complexity in both computation and memory. To achieve truly efficient, real-time segmentation, we believe it is crucial to combine:
\begin{itemize}
    \item \emph{Lightweight convolutions,} which preserve spatial detail while limiting computational load.
    \item \emph{Memory-efficient self-attention,} which expands the receptive field and captures global dependencies without exploding memory usage.
\end{itemize}

In this paper, we propose a \textbf{Hybrid Convolutional-Transformer Architecture Search (HyCTAS)} framework to \emph{automatically} discover powerful yet efficient segmentation networks. Rather than performing a manual, trial-and-error insertion of self-attention layers into a high-resolution CNN, HyCTAS leverages a multi-branch supernet search space and \textcolor{black}{an evolutionary computation-based multi-objective optimization strategy (NSGA-II)}. By jointly optimizing for latency and accuracy (mIoU), HyCTAS identifies an optimal “mix” of convolution and attention at multiple scales and resolutions. As a result, we obtain a family of architectures that lie on the \textcolor{black}{approximate Pareto front} of speed and performance.

\textbf{Our key contributions can be summarized as:}
\begin{itemize}
    \item \textbf{Multi-branch search framework}: We encode parallel CNN branches at multiple resolutions, each of which may contain either lightweight convolution or memory-efficient attention. This unified supernet preserves high-resolution feature streams while optimizing for real-time inference.
    \item \textbf{Flexible multi-objective optimization}: We adopt \textcolor{black}{an evolutionary computation-based approach (NSGA-II)} to search for efficient architectures that yield a range of trade-offs between speed (latency) and segmentation accuracy (mIoU). \textcolor{black}{We explicitly return multiple candidates from the non-dominated set in a single search.}
    \item \textbf{Practical contribution focus}: \textcolor{black}{The contribution lies primarily in the \emph{NAS strategy and search space design} for hybrid high-resolution networks rather than in new fusion operators; the lightweight convolution and memory-efficient attention are modest adaptations chosen to make the search space effective and hardware-friendly.}
    \item \textbf{Strong performance on multiple tasks}: Extensive experiments on benchmarks such as Cityscapes, ADE20K, and COCO demonstrate that our searched models consistently outperform prior works in semantic and panoptic segmentation—while significantly reducing computational cost and increasing inference speed.
\end{itemize}

\section{Related Work}
\label{sec:related}

\subsection{Efficient Networks for Image Segmentation}
Semantic segmentation has long benefited from fully convolutional networks~\cite{long2015fully}, which effectively adapt classification backbones to dense prediction tasks. Subsequent refinements, such as atrous/dilated convolutions in DeepLab~\cite{chen2017deeplab} and multi-branch high-resolution designs in HRNet~\cite{WangSCJDZLMTWLX19}, steadily pushed performance. Meanwhile, real-time or low-complexity methods such as ICNet~\cite{zhao2018icnet}, BiSeNet~\cite{yu2018bisenet}, and Fast-SCNN~\cite{poudel2019fast} introduced architectural optimizations (e.g., factorized convolutions, multi-resolution branches) to reduce latency. In more recent works, lightweight backbone designs~\cite{howard2019searching, wan2023seaformer} target embedded devices and further illustrate the trade-off between efficiency and accuracy. Among recent lightweight designs, LRDNet adopts a compact encoder–decoder with a refined dual-attention module to enable real-time segmentation~\cite{zhuang2021lrdnet}.

\textcolor{black}{Recent hybrid CNN–Transformer segmenters continue to improve efficiency by combining local and global modeling. Examples include SegFormer~\cite{xie2021segformer} and RTLinearFormer~\cite{gu2025rtlinearformer}, which emphasize lightweight attention for mobile/real-time settings. Our work differs by \emph{searching} the hybrid high-resolution design rather than handcrafting a single architecture. Extensive surveys~\cite{poyser_neural_2024} and the emergence of recent lightweight and hybrid architectures~\cite{wan2023seaformer,xie2021segformer} underscore the ongoing advancements since 2020.}

\textcolor{black}{Recent advances in efficient segmentation include HSNet-L~\cite{zhang2022hsnet}, which uses hierarchical similarity networks for few-shot segmentation; AFFormer-B~\cite{dong2023head}, which employs adaptive frequency Fourier transforms for efficient attention; and Vim-S~\cite{zhu2024vision}, which leverages bidirectional state space models for linear-time sequence modeling. These methods demonstrate diverse approaches to balancing accuracy and efficiency in modern segmentation tasks.}

Unlike most prior approaches that \emph{manually} reduce resolution or remove branches to save compute, our HyCTAS approach \emph{searches} across multi-resolution sub-networks, identifying the best structure automatically. This not only exploits the advantages of high-resolution representations but also finds the optimal placement of memory-efficient self-attention modules.

\subsection{Neural Architecture Search}
Neural Architecture Search (NAS) has emerged to automate network design~\cite{Tan_2019_CVPR, baymurzina2022review}. Early NAS solutions often employed reinforcement learning or evolutionary algorithms to explore cell-based search spaces~\cite{liu2018darts}. Although these methods can discover effective designs, their applicability to large-scale semantic segmentation has been limited by high computational overhead.

Recent works have extended NAS to semantic segmentation~\cite{zhang2019customizable, liu2019auto, chen2020fasterseg, li2020learning}, but many still rely on a single-path or single-resolution paradigm. \textcolor{black}{Comprehensive surveys~\cite{poyser_neural_2024} and recent lightweight and hybrid designs~\cite{wan2023seaformer,xie2021segformer} highlight continuing progress beyond 2020.} Our HyCTAS goes further by accommodating \emph{multiple branches} (with varying spatial resolutions) within one unified supernet. Moreover, our multi-objective search can discover a \emph{set} of \textcolor{black}{non-dominated} architectures—each balancing accuracy and speed differently—in a single pass.

\textcolor{black}{Multi-objective NAS.} Beyond single-objective accuracy, multi-objective formulations jointly consider latency/FLOPs/parameters. We adopt NSGA-II~\cite{deb2002fast} for its strong performance and simplicity, and we follow the terminology recommended in recent discussions of multi-objective NAS benchmarking (e.g., reporting an \emph{approximation front}/\emph{non-dominated set} rather than claiming true Pareto optimality on vast spaces). \textcolor{black}{See \cite{deb2002fast} and the broader benchmarking perspective summarized in contemporary surveys~\cite{poyser_neural_2024}.}

\subsection{Transformer in Vision Tasks}
Transformers~\cite{vaswani2017attention} originally showed remarkable success in NLP, and have recently gained popularity in vision tasks. Fully attentional backbones like Vision Transformer~\cite{dosovitskiy2020} and SETR~\cite{SETR} capture global context effectively but incur high memory costs, especially for large input resolutions. Hybrid strategies like DETR~\cite{carion2020end} combine CNNs with transformers for detection and segmentation. An efficient hybrid example is RTLinearFormer, which couples linear attentions with a dual-resolution backbone to reach real-time speed on urban-scene datasets~\cite{gu2025rtlinearformer}.  

\textcolor{black}{
Transformers excel at modeling long-range dependencies but incur quadratic cost in memory and computation at high resolutions~\cite{vaswani2017attention,dosovitskiy2020,SETR}. We therefore blend lightweight convolutions with a \emph{memory-efficient self-attention} module and let the search decide \emph{where/when} to apply attention across scales, striking a better accuracy–runtime balance than manual designs.
}

\section{Proposed Approach}\textcolor{black}{}\label{sec:method}
\subsection{Overall Workflow}
Our HyCTAS framework can be summarized in four main steps:

\begin{enumerate}
    \item \textbf{Supernet Construction}: We construct a multi-branch supernet inspired by HRNet, preserving parallel streams at various resolutions (e.g., 1/8, 1/16, 1/32). In each layer, we introduce \emph{searchable cells} (for choosing lightweight convolution or memory-efficient self-attention) and \emph{searchable nodes} (for fusion and skip connections).
    \item \textbf{Supernet Training}: We randomly sample sub-networks from the supernet during training. To ensure coverage of multi-resolution paths, we bias the sampling probabilities so that deeper or wider sub-networks still appear sufficiently often.
    \item \textbf{Multi-Objective Search}: Once the supernet weights are trained, we evaluate candidate sub-networks by measuring both segmentation accuracy (mIoU) and latency. We formulate a multi-objective problem and adopt an evolutionary algorithm (NSGA-II) to efficiently traverse the architecture space.
    \item \textbf{Architecture Selection}: The search yields a set of \textcolor{black}{non-dominated (approximate Pareto-optimal)} architectures, each offering a distinct trade-off (e.g., ultra-fast inference vs. higher accuracy). We finally retrain the chosen models from scratch to confirm their performance.
\end{enumerate}

\textcolor{black}{For clarity, we use multi-branch to denote parallel streams operating at different resolutions that are maintained throughout the network, and multi-objective to denote our multi-objective optimization over both accuracy (mIoU) and latency. The search returns an \emph{approximate Pareto front (non-dominated set)} of architectures.}

\begin{figure*}[t]
	\centering
	\includegraphics[width=1.0\textwidth]{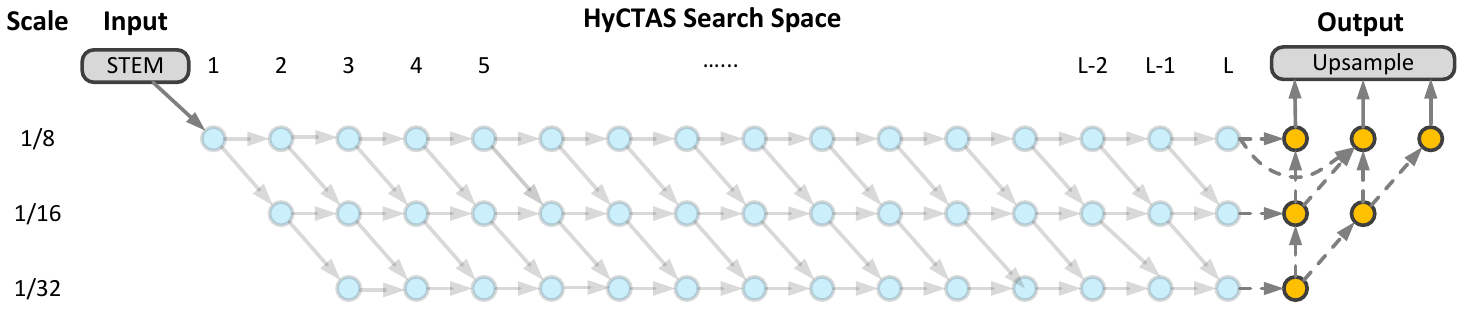}
	\caption{An illustration of our HyCTAS search space. It preserves the multi-resolution and multi-branch properties of HRNet by introducing two types of searchable components: cells and nodes, which extend HRNet with multi-module fusion ability. \textcolor{black}{Blue circles denote cells (operator selection; lightweight conv or memory-efficient self-attention), yellow circles denote nodes (fusion/skip gates controlling inter-branch connections). Solid arrows indicate active connections for a sampled sub-network; dashed arrows indicate optional connections available in the supernet that may or may not be selected during search. Upsampled features are concatenated and fused by a $3\times3$ conv.}}
	\label{fig:intro}
\end{figure*}

\subsection{Revisiting HRNet Architecture}
Learning high-resolution representations is crucial for good performance in image segmentation. HRNet achieves this by maintaining \textbf{multi-branch} convolution streams in parallel to preserve high-resolution representations throughout the whole network and generates reliable high-resolution representations with strong position sensitivity through repeatedly fusing the representations from \textbf{multi-resolution} streams.

In order to preserve these two key properties, we design our HyCTAS supernet to enable searching for variant branch numbers, depths, widths, and spatial resolutions all in one. As shown in Figure \ref{fig:intro}, our supernet contains two types of searchable components: Cells and Nodes.

Given a supernet with $L$ layers, after an input image goes through the initial stem layers, the feature resolutions are downsampled to strides $S = \{1/8, 1/16, 1/32\}$ and fed into multiple searchable cells $C = \{C^1_{s}, C^2_{s}, \dots, C^L_{s}\}$. Then the features are fused into searchable nodes $N = \{N^1_{s}, N^2_{s}, \dots, N^{L-1}_{s}\}$ consecutively, where the cells and nodes are alternately stacked. The searching domain is different for cells and nodes. A cell could contain multiple modules such as convolution, transformer, and so on, which extends the network to be capable of combining \textbf{multi-module} into a sub-network. Meanwhile, a node is dedicated to fusing multiple branches from different resolutions together and behaves as a binary gate.

\textcolor{black}{We use three scales (1/8, 1/16, 1/32) rather than four as in HRNet because keeping a 1/4 branch with attention significantly increases memory and latency for high-resolution inputs. Instead, we preserve 1/4 information in the stem and early fusion, and we allow the search to place attention at shallower stages when operating with fewer branches. This choice improves search tractability and real-time inference while retaining accuracy.}

\textcolor{black}{We distinguish between cells and nodes. In our implementation, a \emph{cell} selects an operator (lightweight conv or memory-efficient self-attention) and its channel multiplier, whereas a \emph{node} decides whether to (i) pass-through, (ii) fuse with upsampled/downsampled sibling features via concatenation+$3\times3$ conv, or (iii) drop the incoming edge. This separation of domains keeps operator selection distinct from topology/fusion decisions.}

Since our architecture needs to support arbitrary branch fusion, the outputs of each architecture are not fixed, \textit{i.e.}, the number of the last feature map channels and the branch position of the final outputs are not fixed. To solve these problems, we propose a feature aggregation head module that can adapt to arbitrary outputs of different architectures. The feature map of shape ($C_{*}\times H\times W$) is first adjusted in channels by a $1\times1$ Slimmable Convolution layer \cite{yu2018slimmable} to the shape of ($C_{s}\times H\times W$). If the feature map size is not the max resolution in the search space, it will be bilinearly upsampled to match the shape of the other feature map ($C_s \times 2H \times 2W$). Then, the two feature maps are concatenated and fused together with a $3\times3$ convolution layer. We enumerate all the output combinations and set a head for each type of output $F_{\theta}$, so there are a total of six different heads in our search space, \textit{i.e.}, $\theta \in \{(8), (16), (32), (8,16), (8,32), (8,16,32)\}$.

Unlike previous works \cite{liu2019auto, chen2020fasterseg} that limit the network search to one path, our HyCTAS supernet is capable of modeling sub-networks with arbitrary numbers of branches from different resolutions, which better captures the spirit of HRNet.

\textcolor{black}{Regarding the search space, cells choose between lightweight convolution and memory-efficient self-attention (Section~\ref{sec:ops}), while nodes decide whether and how to fuse incoming branches at each resolution. The branch count is variable (1--3), and depths/widths per branch are adjustable. This compositional design yields a large combinatorial space yet remains structured enough for efficient search.}

\subsection{Combining Multiple Modules}\label{sec:ops}
Convolution facilitates the learning of visual features. With the weight-sharing mechanism, the features extracted from a convolutional layer are translation invariant. At the same time, the features are also locality-sensitive, where every operation only takes into account a local region of the image. Thus, convolutional operations fail to encode relative spatial information and result in a limited receptive field.

However, large receptive fields are often required to track long-range dependencies within an image, which in practice involves using large kernels or long sequences of convolutional layers at the cost of losing efficiency. Recent works have presented the techniques of Transformers \cite{vaswani2017attention} to overcome the limitations presented by convolutional operations in an efficient way.

Transformers use the concept of multi-head self-attention. Such an attention mechanism models the importance between each pixel within the whole input, which results in a maximized receptive field. Compared with convolutional operations, transformers also have the problem of large computational overhead, especially when the feature map size is very large, as its computation and memory usage are quadratic to the input size.

Therefore, both operations do not meet our goal of pursuing efficiency if naively incorporated in our multi-resolution multi-branch framework for high-resolution image segmentation learning. Hence, we propose to use a memory-efficient self-attention module and a lightweight convolution module.

\begin{figure}[t]
	\hspace{0.5cm}
		\centering
		\includegraphics[width=0.5\textwidth]{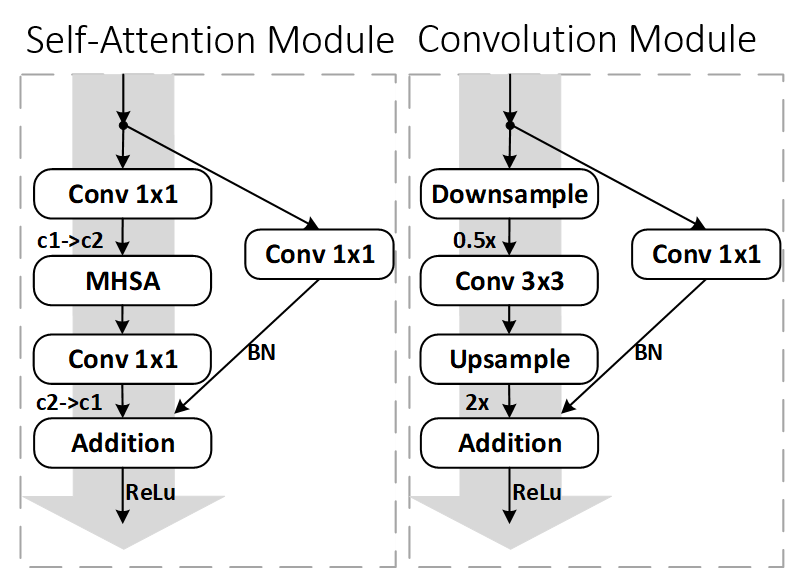}
		\caption{An illustration of our searchable modules. We design a memory-efficient self-attention module and a lightweight convolution module and search for the best combination.}
		\label{fig:op}	
	\vspace{0.cm}
\end{figure}

\begin{table*}[!t]
	\small
	\begin{center}
		\caption{The difference between a typical operation and the proposed efficient operation. FLOPs are measured using an input of size 1$\times$256$\times$32$\times$64. The receptive field (RF) is relatively compared with the standard convolution.}
			\begin{tabular}{cccccc}
				\toprule
				Operation Type & FLOPs (G) & Params (M) & RF & Scale-to-Input\\ \midrule
				Convolution & $1.2$ & $0.59$ & $1$ & Linear\\
				\textcolor{black}{Lightweight Convolution} & $0.4$ & $0.66$ & $2$ & Linear\\
				Transformer & 4.3 & 12.62 & $Max$ & Quadratic\\
				Memory-efficient Self-attention & $0.6$ & $0.81$ & $Max$ & Quadratic\\ \bottomrule
			\end{tabular}\label{table:ops}
	\end{center}
\end{table*}
\textcolor{black}{Why can the lightweight convolution have slightly more parameters than vanilla conv in Table~\ref{table:ops}? The parallel $1\times1$ branch used to preserve high-resolution details adds a small number of parameters; in return, it reduces information loss from resizing and improves accuracy at similar or lower FLOPs. Our search can also choose it less frequently when strict parameter budgets are prioritized.}

\textbf{Memory-efficient Self-attention Module}: Instead of directly borrowing typical transformer operations like previous works \cite{SETR, OCR}, we only employ the self-attention layer \cite{vaswani2017attention} without the \textcolor{black}{feed-forward network (FFN)} to capture the global context while maximizing efficiency. We then use $1 \times 1$ convolutions for reducing and then increasing dimensions, leaving the self-attention module as a bottleneck with smaller than input/output channels. This largely reduces the memory usage required in the self-attention layer. Meanwhile, we add an extra bypassed $1 \times 1$ convolution module for preserving uncompressed residue. Our memory-efficient self-attention module operation can utilize the largest receptive field while keeping relatively small FLOPs and memory footprint.

\textcolor{black}{In relation to linear/efficient attention, our module keeps exact (quadratic) dot-product attention but applies an aggressive channel bottleneck around attention to lower the memory/time cost. This differs from kernelized or low-rank "linear" attention variants that approximate attention to achieve $\mathcal{O}(N)$ complexity; we chose exact attention with a bottleneck to maintain accuracy while controlling cost in high-resolution settings.}

\textbf{Lightweight Convolution Module}: To maintain a large receptive field while minimizing computational costs, methods like Zoomed Convolution \cite{chen2020fasterseg} and Dilated Convolution \cite{chen2017deeplab} employ bilinear sampling for input feature map resizing, inevitably leading to information loss. Addressing this, we introduce a novel, efficient convolution that incorporates a parallel $1 \times 1$ convolution to preserve high-resolution details, as shown in Figure~\ref{fig:op}. This approach connects the input feature map directly to the zoomed convolution's output, effectively maintaining input resolution. Our lightweight convolution module significantly reduces information loss with minimal parameter increase, ensuring both speed and accuracy.

\textcolor{black}{We clarify that we do not claim the above modules as architectural novelties per se; rather, they are pragmatic choices that keep the search space expressive yet tractable in memory and latency, aligning with the main contribution of HyCTAS (the multi-objective NAS over multi-branch high-resolution backbones).}

The detailed measurements of our memory-efficient self-attention module and lightweight convolution module are shown in Table~\textcolor{black}{\ref{table:ops}}. Properly combining the convolutional module and the self-attention module plays a significant role in improving model performance and efficiency. Hence, we further deploy a search algorithm to solve this problem.

\subsection{Multi-objective Search}
Inspired by \cite{randomNAS}, we formulate the search for the optimal architecture $\alpha^{*}$ as a two-stage optimization problem. The architecture search space $\mathcal{A}$ is encoded in the supernet, denoted as $\mathcal{N}(\mathcal{A}, \mathcal{W})$. Each candidate architecture $\alpha \in \mathcal{A}$ shares its weights across $\mathcal{W}$. In the first stage, we optimize the supernet weights $\mathcal{W}$ by:
\begin{equation}\label{equ:supernet}
\mathcal{W} =  \mathop { \mathrm{arg \, min} }\limits_{\mathcal{W}} \mathcal{L}_{train}(\mathcal{N}(\mathcal{A}, \mathcal{W}))
\end{equation}
where $\mathcal{L}_{train}$ denotes the training loss.

The prior distribution of $\Gamma(\mathcal{A})$ is important. In order to meet our goal of searching for architectures with arbitrary branches from different resolutions, we pre-define this prior in the training process of the supernet. Since the sub-network with more branches contains the sub-network with fewer branches, we empirically set the sub-network with more branches to have a higher sampling probability.
\begin{equation}\label{equ:sampling}
\rho_{i} < \rho_{i+1}, \quad i \in \{1,2,3\}
\end{equation}
where $\rho_{i}$ means the sampling probability of architectures containing $i$ branches. Other components in the sub-network $\alpha$ are sampled from a uniform distribution.

Once the supernet training is finished, the second stage is to search for architectures by ranking the accuracy and latency of each sub-network $\alpha \in \mathcal{A}$. We convert the architecture searching problem to a standard multi-objective optimization (MOP) problem.

The formulation consists of decision variables, objective functions, and constraints. The optimization objectives are as follows:
\begin{align}
\label{1}
\begin{array}{*{20}{l}}
{Maximize\;\;\;y = f(\alpha) = ({f_1}(\alpha),{f_2}(\alpha),...,{f_k}(\alpha))}\\
{S.t.\;\;\;\;\;\;e(x) = ({e_1}(\alpha),{e_2}(\alpha),...,{e_m}(\alpha)) \le 0}\\
{{\rm{among\;\;\;\;x}} = ({\alpha_1},{\alpha_2},...,{\alpha_n}) \in \mathcal{A}}\\
{\;\;\;\;\;\;\;\;\;\;\;\;\;\;\;\;y = ({y_1},{y_2},...,{y_k}) \in Y}
\end{array}
\end{align}

Where $\alpha$ represents the decision vector, \textit{i.e.}, the searched network architecture, $y$ represents the target vector, \textit{i.e.}, the mIoU and latency. $\mathcal{A}$ represents the search space formed by the decision vector $\alpha$, and $Y$ represents the target space formed by the target vector $y$. The constraint \begin{math}e(x)\le0\end{math} determines the feasible value range of the decision vector.
In order to ensure that the searched architectures are effective, we set the following four constraints:
1) The number of paths in each sub-network needs to be greater than 0;
2) For a sub-network that contains multiple paths, the locations of down-sampling should be different;
3) In the down-sampling position, skip connections cannot be selected;
4) The memory usage of a sub-network during training cannot exceed the hardware limit.

\textcolor{black}{Objectives and latency measurement. We use $k{=}2$ objectives: maximize mIoU and minimize inference latency. Latency is measured with batch size 1 using CUDA events~\cite{paszke2019pytorch} at the corresponding deployment resolution on the same GPU (Cityscapes: 1024$\times$2048; COCO: 640$\times$640). We report FPS as $\mathrm{FPS}=1/\mathrm{latency}$ (seconds). This protocol is used consistently during search-time evaluation and for final reporting to ensure apples-to-apples comparisons.}

Different from the single-objective optimization problem, there is usually no unique optimal solution in an MOP problem, but a Pareto optimal solution set. Hence, we adapt the popular evolutionary algorithm NSGA-II \cite{deb2002fast} to solve the optimization problem. The details of our search algorithm can be found in the supplemental material.

\textcolor{black}{We encode candidates one-hot at the cell/node level and use tournament selection, one-point crossover at the branch-level topology, and bit-flip mutation over cell/operator choices with per-variable probability \(1/n_{\text{var}}\), while respecting the constraints above. Non-dominated sorting and crowding distance then guide selection toward an \emph{approximate Pareto front (non-dominated set)}.}

The detailed search algorithm is presented in Algorithm~\ref{alg:search}. One-hot encoding is employed to encode the sub-network $\alpha$. When the sampling is over, we will decode the sub-network to evaluate its corresponding mIoU and latency. We solve this multi-objective optimization (MOP) problem under the NSGA-II~\cite{deb2002fast} framework.
\textcolor{black}{\textbf{NSGA-II configuration.} We use a seeded population of size \(40\) for \(20\) generations. Crossover is one-point at the branch-level topology with probability \(p_c{=}0.9\); mutation is bit-flip over categorical genes with per-variable rate \(1/n_{\text{var}}\). Duplicate architectures are eliminated, and selection uses non-dominated sorting with crowding-distance tournament. Searches are run independently for 1-, 2-, and 3-branch settings, yielding \(40\times20{=}800\) evaluations per setting and \(2400\) in total.}

\begin{algorithm}
    \caption{Multi-objective Search}
    \label{alg:search}
    \begin{algorithmic}[1]
        \Require supernetwork $\mathcal{N}$ with weights $\mathcal{W}$, the number of branches $b$, the number of generations $N$, population size $n$, validation dataset $D_{val}$, \textcolor{black}{latency cap $S_{\max}$, minimum accuracy $\mathrm{mIoU}_{\min}$}
        \Ensure \textcolor{black}{$K$ architectures from the non-dominated set (approximate Pareto front)}
        \State Uniformly generate the populations $P_{0}$ and $Q_{0}$ until each has $n$ individual architectures.
        \For{$i=0$ \textbf{to} $N-1$}
            \State $R_{i} = P_{i} \cup Q_{i}$
            \State \textcolor{black}{$F = \text{Non-Dominated-Sorting}(R_{i})$}\textcolor{black}{\;\;\# fronts and ranks}
            \State Pick $n$ individual architectures to form $P_{i+1}$ according to the ranks and crowding distance.
            \State $Q_{i+1} = \varnothing$
            \While{$\text{size}(Q_{i+1}) < n$}
                \State $M = \text{Tournament-Selection}(P_{i+1})$
                \State $q_{i+1} = \text{Crossover}(M)$
                \State \textcolor{black}{$q_{i+1} \leftarrow \text{Mutate}(q_{i+1})$}
                \If{\textcolor{black}{$\mathrm{Latency}(q_{i+1}) > S_{\max}$}}
                    \State \textbf{continue}
                \EndIf
                \State \textcolor{black}{\(\mathrm{mIoU}(q_{i+1}) \leftarrow \mathrm{Forward}(\mathcal{N}_a;\mathcal{W})\)} \Comment{\textcolor{black}{uses \(\mathcal{W}\)}}
                \State \textcolor{black}{\(\mathrm{Latency}(q_{i+1}) \leftarrow \mathrm{TimeForward}(\mathcal{N}_a;\mathcal{W})\)} \Comment{\textcolor{black}{uses \(\mathcal{W}\)}}
                \If{$\mathrm{mIoU}_{(q_{i+1})} > \mathrm{mIoU}_{\mathrm{min}}$}
                    \State Add $q_{i+1}$ to $Q_{i+1}$
                \EndIf
            \EndWhile
        \EndFor
        \State \textcolor{black}{Let $\mathcal{P}$ be the first front in $F$; return the top-$K$ by crowding distance from $\mathcal{P}$ for retraining.}
    \end{algorithmic}
\end{algorithm}

\textcolor{black}{For retraining, we take the returned $K$ architectures and train them from scratch with the schedule in Section~\ref{sec:exp}; we then report final results on the validation/test sets using these retrained models.}

\section{Experiments}
\label{sec:exp}
We evaluate our method in both semantic segmentation task and the panoptic segmentation tasks. Panoptic segmentation was introduced by~\cite{kirillov2019panoptic} to combine semantic and instance segmentation, in which all object instances are uniquely segmented to make the task even more challenging.

\textcolor{black}{Unless otherwise specified, all results are single-scale, single-crop, and without flipping. For speed, we report latency/FPS with batch size 1 after a warm-up, using CUDA events and averaged over multiple runs on the same hardware noted below to ensure reproducibility.} \textcolor{black}{Unless otherwise specified, Cityscapes latency/FPS is measured at 1024$\times$2048 and COCO latency/FPS is measured at 640$\times$640.}

\begin{figure*}[th]
	\centering
	\vspace{-0.0cm}
	\includegraphics[width=1\linewidth]{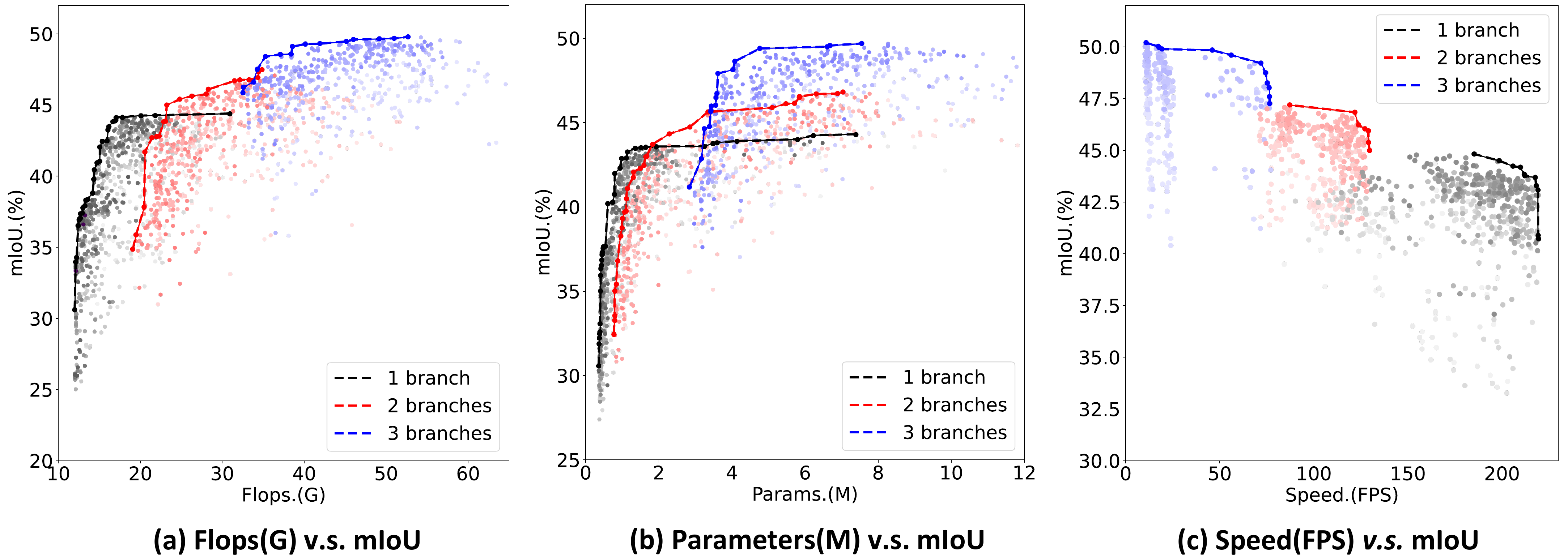}
	\caption{The searched architectures under different objectives on the Cityscapes validation set. Our proposed HyCTAS approach can search for multiple efficient architectures with arbitrary branches in a single run. The color from light to dark indicates that our search results are getting better and better as the search progresses. The solid line represents architectures on the \textcolor{black}{approximate Pareto front (non-dominated set)}. Best viewed in color.}
	\label{fig:arch}
	\vspace{-0.0cm}
\end{figure*}


\subsection{Benchmark and Evaluation Metrics}
\textbf{Cityscapes}~\cite{Cordts2016Cityscapes} is a popular dataset that contains a diverse set of stereo video sequences recorded in street scenes from 50 different cities. It has 5,000 high-quality pixel-level annotations. There are 2,975 images for training, 500 images for validation, and 1,525 images without ground truth for testing to ensure a fair comparison. The dense annotation contains 19 classes for each pixel.

\textbf{COCO}~\cite{lin2014microsoft} is a much larger and more challenging dataset for object detection and segmentation. There are a total of 133 classes, which include 80 'thing' and 53 'stuff' classes. We train our models on the train2017 set (118k images) and evaluate on the val2017 set (5k images).

\textbf{ADE20K}~\cite{zhou2017scene} has 20k images for training, 2k images for validation, and 3k images for testing. It is used in the ImageNet scene parsing challenge 2016 and has 150 classes and diverse scenes with 1,038 image-level annotations.

\textbf{Evaluation Metrics.} The mean intersection over union per class (mIoU), panoptic quality (PQ), and FPS (frames per second) are used as the metrics for semantic segmentation, panoptic segmentation, and speed, respectively.

\begin{figure*}[th]
	\centering
	\vspace{-0.cm}
	\hbox{\includegraphics[width=1\linewidth]{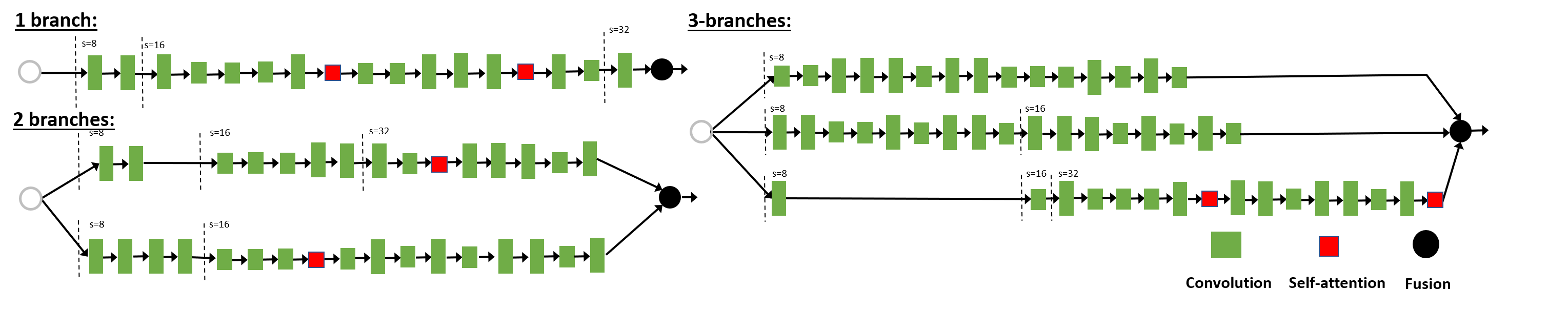}}
	\caption{An illustration of our searched architectures with multiple branches. \textcolor{black}{Green denotes lightweight convolution, red denotes memory-efficient self-attention, and gray indicates skip/fusion nodes.}}
	\vspace{-0.cm}
	\label{fig:network}
\end{figure*}

\subsection{Setup}
\textbf{Search.} During training of the supernet, the sampling ratios for $\rho_{1}$, $\rho_{2}$, and $\rho_{3}$ are set to 0.2, 0.3, and 0.5, respectively.
The initial channel count for the supernet is 64, and the mini-batch size is set to 16. The supernet is trained on the Cityscapes dataset with $256\times512$ images from the training set. We use the SGD optimizer with a momentum of 0.9 and a weight decay of $5\times10^{-4}$ to optimize the supernet weights. The learning rate is set to 0.02 with an exponential learning rate decay of 0.992. The entire supernet optimization takes about 120 hours on a single V100 GPU. 

In the search process, the initial population $P_{0}$ is set to 40 and generated by uniform random sampling. The crossover probability is set to 0.9, and mutation uses bit-flip over categorical genes with a per-variable rate \(1/n_{\text{var}}\).
Before the inference of a sub-network, the statistics of all Batch Normalization operations are recalculated on a random subset of training data (1000 images in Cityscapes).
The number of generations is 20, and we search sub-networks with different branch numbers separately. Hence, a total of 2400 network sub-networks are searched.

\textcolor{black}{Latency protocol. Latency during search and at evaluation is measured with batch size 1 using CUDA events, averaged across 100 forward passes after 50 warm-up passes at the corresponding input resolution on the same GPU model. Specifically, Cityscapes latency/FPS is measured at 1024$\times$2048 and COCO latency/FPS at 640$\times$640. We report FPS as $\text{FPS} = 1/\text{latency}$ (seconds). }

\textbf{Retrain.} 
Similar to~\cite{liu2019auto}, we also use the filter multiplier to control the model capacity, and the default values for HyCTAS-S and HyCTAS-L are set to 12 and 16, respectively. The final output stride of HyCTAS-S is 8, and the other big models' final output strides are set to 4 by merging the low-level features in the stem. We use the SGD optimizer with a momentum of 0.9 and a weight decay of $5\times10^{-4}$ to optimize the searched networks. The 'poly' learning rate policy is employed with an initial learning rate of 0.005. Data augmentations on all datasets, \textit{i.e.}, random crop, flipping, etc., are consistent with \cite{cheng2020panoptic}. On the Cityscapes dataset, the training images are cropped to $512\times1024$ and the mini-batch size is set to 32. It is worth noting that all our results are trained from scratch without ImageNet pretraining, and we only use the training set without any extra data.

\textbf{Evaluation.} We use an Nvidia GeForce RTX 3090 and an Intel Xeon E5-2680-v4 CPU (2.40GHz) for benchmarking the computation cost. When testing the inference speed, an image with a batch size of 1 is first loaded into the GPU memory and the model is warmed up to reach a steady speed. After that, the inference time is measured by running the model for six seconds. All the reported results, unless specified, are evaluated without multi-scale inference and flipping. The mean intersection over union per class (mIoU), panoptic quality (PQ) are used as the evaluation metrics for semantic segmentation and panoptic segmentation, respectively.

\textcolor{black}{\textbf{Baseline reproduction and pretraining.} We re-ran from scratch (no ImageNet pretraining) a subset of baselines under our protocol and hardware, including FCN/PSPNet/DeepLabV3+ (MobileNetV2) and HRNetV2-W18S. For methods commonly reported with ImageNet pretraining (e.g., DeepLabV3+ with ResNet-101, SegFormer), we cite their original numbers and mark them as borrowed where applicable; we do not mix pretraining regimes in our claims and explicitly state the training regime for HyCTAS (from scratch) in all tables/sections.}

\subsection{Model Exploration}
We begin our experiments with the exploration of our searched architectures. As mentioned in the previous section, our search framework is capable of optimizing towards multiple different objectives. To better understand the architecture distributions under different objectives, we conduct three different searches, which cover the objectives "FLOPs vs. mIoU," "Parameters vs. mIoU," and "Speed vs. mIoU," respectively. As shown in Figure \ref{fig:arch}, we visualize the evolution processes for different objectives. As our method is capable of searching sub-networks with an arbitrary number of branches, we visualize the group of architectures with different numbers of branches in different colors. We also plot the Pareto boundary at the end of the search. It is interesting to find out that searching for different objectives largely affects the architecture distributions. For the following experiments, we choose architectures on the Pareto boundary from searching under "Speed vs. mIoU" because of better model separation and closer alignment to real-world requirements.

\textcolor{black}{To avoid confusion, we refer to the plotted envelope as the \emph{approximate Pareto front (non-dominated set)}, estimated from the finite set of candidates evaluated during search.}

We also visualize the searched architectures with multiple branches in Figure \ref{fig:network}. It is interesting to observe how the self-attention module is placed under a different number of branches. When the number of branches is 1, the self-attention modules can be placed in higher resolution to increase the network's performance; when the number of branches is 3, the self-attention modules would be placed at a deeper stage to better fuse the features from multiple branches. This phenomenon well aligns with self-attention's advantages and limitations.



\subsection{Comparison to SOTA Models}

\begin{table*}[th]
	\vspace{-0.cm}
	\caption{\textcolor{black}{Comparison of semantic segmentation on Cityscapes. $\dagger$: speed on NVIDIA Xavier; $\sharp$: speed with TensorRT acceleration; $\star$: trained from scratch. Entries without $\dagger/\sharp$ are measured by us on RTX 3090 at 1024$\times$2048 with batch size 1. “-” denotes values not reported by the original papers. \textbf{Pretrain} indicates ImageNet pretraining status (Y/N/-).}}
	\centering
	\small
	\begin{tabular}{c|c|cc|ccc|c} 
			\toprule[1.5pt]
			\multirow{2}{*}{Methods} & \multirow{2}{*}{Input Size} & \multicolumn{2}{c|}{\textcolor{black}{mIoU (\%) $\uparrow$}} & FLOPs $\downarrow$ & Params $\downarrow$ & Speed $\uparrow$ & \multirow{2}{*}{ \textcolor{black}{Pretrain}} \\
			&  & val & test & (G) & (M) & (FPS) &  \\
			\midrule[1.5pt]
			ICNet~\cite{zhao2018icnet}& 1024x2048 & 67.7 & 69.5 & - & - & 37.7 & \textcolor{black}{-} \\
			BiSeNet~\cite{yu2018bisenet}& 768x1536 & 69.0 & 68.4 & - & - & 105.8  & \textcolor{black}{Y} \\
			Fast-SCNN~\cite{poudel2019fast}& 1024x2048 & 68.6 & 68.0 & - & 11.1 & 123.5  & \textcolor{black}{N} \\
			DF1-Seg-d8~\cite{li2019partial}& 1024x2048 & 72.4 & 71.4  & - & - & 136.9  & \textcolor{black}{Y} \\
			CAS~\cite{zhang2019customizable}& 768x1536 & 71.6 & 70.5 & - & - & 108.0 & \textcolor{black}{Y} \\
			MobilenetV3~\cite{howard2019searching}& - & 72.4 & 72.6  & 9.7 & 1.5 & $10.8^\dagger$ & \textcolor{black}{Y} \\
			SqueezeNAS~\cite{shaw2019squeezenas}& - & 73.6 & 72.5 & 19.6 & 1.9 & $10.2^\dagger$& \textcolor{black}{Y}\\
			FasterSeg~\cite{chen2020fasterseg}& 1024x2048 & 73.1 & 71.5 & 28.2 & 4.4 & $149^{\sharp}$ & \textcolor{black}{-} \\
   			SeaFormer-S~\cite{wan2023seaformer}& 1024x2048 & 70.7 & 71.0 & 2.0 & - & 7.7 & \textcolor{black}{Y} \\
   			SeaFormer-B~\cite{wan2023seaformer}& 1024x2048 & 72.2 & 72.5 & 3.4 & - & 4.8 & \textcolor{black}{Y} \\
			$\text{Dynamic}_{L32}$~\cite{li2020learning}& 1024x2048 & 79.2 & 80.0 & 242.3 & - & -& \textcolor{black}{Y}\\ 
			$\text{SETR}_{PUP}$~\cite{SETR} & 768x768 & 79.3 & - & 318.3 & - & -& \textcolor{black}{Y}\\
			$\text{Auto-DeepLab}_{L}^\star$~\cite{liu2019auto}& 1025x2049 &80.3 &80.4 & 695.0 & 44.4 & 0.5 & \textcolor{black}{N} \\
			$\text{HRNetV2}_{W18S}$~\cite{WangSCJDZLMTWLX19}& 1024x2048 & 70.3 & - & 31.1 & 1.5 & - & \textcolor{black}{Y} \\
			$\text{HRNetV2}_{W40}$~\cite{WangSCJDZLMTWLX19}& 1024x2048 & 80.2 & 80.5 & 493.2 & 45.2 & - & \textcolor{black}{Y} \\
			$\text{OCR}_{HRNetV2-W48}$~\cite{OCR} & 1024x2048 & 80.6 & - & 617.2 & 63.8 & 4.6& \textcolor{black}{Y}\\
            HSNet-L~\cite{zhang2022hsnet} & 512x1024 & 80.0 & - & 117.7 & 33.0 & 51.3 & \textcolor{black}{-}\\
            RTLinearFormer~\cite{gu2025rtlinearformer} & 1024x2048 & 78.4 & - & 117.7 & 24.0 & 66.7 & \textcolor{black}{Y}\\
    		\midrule
			\midrule
			\textbf{$\text{HyCTAS-S}^\star$}& 1024x2048 & 72.3 & 73.0 & 13.0 & \textbf{1.5} & \textbf{159.2} & \textcolor{black}{N} \\
			\textbf{$\text{HyCTAS-M}^\star$}& 1024x2048 & 77.1 & 75.7 & 29.4 & 2.4 & 87.8 & \textcolor{black}{N} \\
			\textbf{$\text{HyCTAS-L}^\star$}& 1024x2048 & \textbf{80.9} & 79.4 & 527.0 & 43.2 & 12.4 & \textcolor{black}{N} \\
			\bottomrule[1.5pt]
	\end{tabular} 
	\label{tab:sota}
\end{table*}

\textcolor{black}{\textbf{Reporting protocol and apples-to-apples.} To avoid mixing measurement regimes, entries without $\dagger/\sharp$ in Table~\ref{tab:sota} are our re-runs on RTX 3090 at 1024$\times$2048, batch size 1. Borrowed speeds marked by $\dagger/\sharp$ retain the original papers' hardware/protocols and are noted for transparency rather than for apples-to-apples claims. We therefore restrict comparative speed statements to methods we re-ran under the same resolution and hardware. We also normalized model naming across tables (e.g., removing “(Ours)” to maintain consistency).}

\begin{table}[!t]
	\centering
	\caption{\textcolor{black}{Comparisons of panoptic segmentation on Cityscapes validation set. $\dagger$: speed on Tesla V100-SXM2 GPU, $\ddagger$: multi-scale + flip evaluation, $\star$: trained from scratch. For entries marked with $\ddagger$, the single fixed input size is not applicable; we denote Input Size as "MS+Flip (var)" to avoid confusion. “-” denotes values not reported by the original papers. \textbf{Pretrain} indicates ImageNet pretraining status (Y/N/-).}}
		\begin{tabular}{c | c | c | c | c | c }
			\toprule[1.5pt]
			\multirow{2}{*}{Methods} & \multirow{2}{*}{Input Size}  & PQ & AP & mIoU & \multirow{2}{*}{ \textcolor{black}{Pretrain}} \\
			& &  (\%) & (\%) & (\%) &  \\
			\toprule[1.5pt]
			$\text{TASCNet}^{\ddagger}$~\cite{li2018learning}& - & 60.4 & 39.1 & 78.7 & \textcolor{black}{Y} \\
			Panoptic FPN~\cite{kirillov2019panoptic}& -  & 58.1 & 33.0 & 75.7 & \textcolor{black}{Y} \\
			AUNet~\cite{li2019attention}& -  & 59.0 & 34.4 & 75.6 & \textcolor{black}{Y} \\
			UPSNet~\cite{xiong2019upsnet}& 1024x2048  & 59.3 & 33.3 & 75.2 & \textcolor{black}{Y} \\
			$\text{UPSNet}^{\ddagger}$~\cite{xiong2019upsnet}& -  & 60.1 & 33.3 & 76.8 & \textcolor{black}{Y} \\
			Seamless~\cite{porzi2019seamless}& 1025x2049  & 60.3 & 33.6 & 77.5 & \textcolor{black}{Y} \\
			AdaptIS~\cite{sofiiuk2019adaptis}& 1024x2048  & 62.0 & 36.3 & 79.2 & \textcolor{black}{Y} \\
			DeeperLab~\cite{yang2019deeperlab}& 1025x2049  & 56.5 & - & - & \textcolor{black}{Y} \\
			$\text{SSAP}^{\ddagger}$~\cite{gao2019ssap}& -  & 61.1 & 37.3 & - & \textcolor{black}{Y} \\
			\midrule\midrule
			$\text{Panoptic-DeepLab}_{HRNet-W48}$ & 1025x2049  & 63.3 & 35.9 & 80.3 & \textcolor{black}{Y} \\
			$\text{Panoptic-DeepLab}_{Xception-71}$ & 1025x2049  & 63.0 & 35.3 & 80.5 & \textcolor{black}{Y} \\
			\textbf{$\text{HyCTAS-L}^\star$} & 1024x2048 & \textbf{63.3} & 34.0 & \textbf{81.3} & \textcolor{black}{N} \\
			\midrule\midrule
			$\text{Panoptic-DeepLab}_{Xception-71}^{\ddagger}$ & MS+Flip (var)  & 64.1 & 38.5 & 81.5 & \textcolor{black}{Y} \\
			\textbf{$\text{HyCTAS-L}^{\ddagger\star}$} & MS+Flip (var)  & \textbf{64.8} & 36.8 & \textbf{81.8} & \textcolor{black}{N} \\
			\bottomrule[1.5pt]
		\end{tabular}
	\label{tab:city_pano}
\end{table}

We compare our method with state-of-the-art methods on two popular benchmarks. We can see that when the speed is more than 100 FPS, our model establishes new state-of-the-arts. HyCTAS-M consistently outperforms recent BiSeNet~\cite{yu2018bisenet}, Fast-SCNN~\cite{poudel2019fast}, and DF1-Seg-d8~\cite{li2019partial}. In particular, our HyCTAS-M achieves 77.1\% mIoU on the Cityscapes validation set with 29.4G FLOPs and 2.4M parameters, surpassing HRNetV2-W18-Small-v1 by 6.8\% with a similar model size. On the Cityscapes test set, our HyCTAS-M surpasses FasterSeg~\cite{chen2020fasterseg} by 4.2\%.
Moreover, the speed of our model reaches an astonishing 87.8 FPS without TensorRT acceleration. Meanwhile, our HyCTAS-S has the smallest FLOPs and parameters and achieves the fastest inference speed at 159.2 FPS. This proves that our models are very competitive when deployed on real hardware.

Without any pretraining or knowledge distillation, our big model HyCTAS-L achieves 80.9 mIoU on the Cityscapes validation set, which is 0.6\% higher than Auto-DeepLab~\cite{liu2019auto}, while using fewer FLOPs. Compared to the expert-designed HRNetV2-W40~\cite{WangSCJDZLMTWLX19}, HyCTAS-L is still 0.7\% ahead on the validation set with nearly 100G fewer FLOPs. Notably, the HyCTAS-L can still run at 12.4 FPS. Since we target fast semantic segmentation, we did not use dilated separable convolution for fast inference speed. Besides, the mIoU is measured without any evaluation augmentation like flipping or multi-scale.

Compared with recent NAS-related models, our method also has obvious advantages in search time and GPU memory overhead: 1) Thanks to our multi-objective searching strategy, our HyCTAS only takes 5 GPU days to search a series of models which can satisfy different demands, \textit{i.e.}, low latency model or high-performance model. By contrast, other NAS methods can only search for one model at a time, so their search overhead will increase linearly as the number of searched models increases. 2) Our model also achieves better performance than the baseline NAS methods. For example, HyCTAS-M is 3.8\% superior to FasterSeg~\cite{chen2020fasterseg} on the Cityscapes validation set.

\textcolor{black}{Our HyCTAS-L achieves superior performance compared to recent transformer-based methods such as HSNet-L~\cite{zhang2022hsnet}, which uses hierarchical similarity networks for few-shot segmentation, and RTLinearFormer~\cite{gu2025rtlinearformer}, which employs linear attention mechanisms for real-time performance. These comparisons demonstrate that our search-based approach can discover more effective hybrid architectures than handcrafted designs.}

Compared with the latest development of Transformer-like models~\cite{SETR, OCR}, our HyCTAS-L model can afford inference using a much larger image and requires fewer FLOPs respectively. This demonstrates the overall advantages of our architectures, which can better combine convolution and self-attention to maximize efficiency.

We also evaluate HyCTAS-L on the Cityscapes validation set with the original image resolution of $1024\times2048$. In Table ~\ref{tab:city_pano}, we see the superior PQ and speed of our HyCTAS-L. Without any inference tricks, our HyCTAS-L achieves 63.3\% PQ and 81.3\% mIoU, which is comparable to Panoptic-Deeplab with the HRNet-W48~\cite{WangSCJDZLMTWLX19} backbone. However, our HyCTAS-L runs twice as fast as Panoptic-Deeplab with the HRNet-W48 backbone. When we use inference data augmentation, our method achieved the best results. Specifically, our HyCTAS-L surpasses Panoptic-Deeplab~\cite{cheng2020panoptic} by 0.7\% on PQ and 0.3\% on mIoU. These results are achieved with only Cityscapes fine-annotated images, without using any extra data (coarse-annotated images, COCO, ImageNet, etc.).

\begin{table}[!t]
	\vspace{-0.cm}
	\centering
	\caption{\textcolor{black}{Comparison of semantic segmentation on ADE20K. “-” denotes values not reported by the original papers. \textbf{Pretrain} indicates ImageNet pretraining status (Y/N/-).}}
	\small
	\begin{tabular}{c|c|ccc|c|c}
			\toprule[1.5pt]
			\multirow{2}{*}{Methods} & \multirow{2}{*}{Encoder} & FLOPs  & Params & Speed & mIoU & \multirow{2}{*}{ \textcolor{black}{Pretrain}} \\
			& & {(G)}  & {(M)} & {(FPS)} & (\%) &  \\
			\midrule[1.5pt]
			FCN~\cite{long2015fully} & MobileNetV2 & 39.6 & 9.8 & 64.4 & 19.7 & \textcolor{black}{Y} \\
			PSPNet~\cite{zhao2017pyramid} & MobileNetV2 & 52.9 & 13.7 & 57.7 & 29.6 & \textcolor{black}{Y} \\
			DeepLabV3+~\cite{chen2018encoder} & MobileNetV2 & 69.4 & 15.4 & 43.1 & 34.0 & \textcolor{black}{Y} \\
			SegFormer~\cite{xie2021segformer} & MiT-B0 & 8.4 & 3.8 & 50.5 & 37.4 & \textcolor{black}{Y} \\
			FCN~\cite{long2015fully} & ResNet-101 & 275.7 & 68.6 & 14.8 & 41.4 & \textcolor{black}{Y} \\
			EncNet~\cite{zhang2018context} & ResNet-101 & 218.8 & 55.1 & 14.9 & 44.7 & \textcolor{black}{Y} \\
			PSPNet~\cite{zhao2017pyramid} & ResNet-101 & 256.4 & 68.1 & 15.3 & 44.4 & \textcolor{black}{Y} \\
			DeepLabV3+~\cite{chen2018encoder} & ResNet-101 & 255.1 & 62.7 & 14.1 & 44.1 & \textcolor{black}{Y} \\
			Auto-DeepLab~\cite{liu2019auto} & NAS-F48-ASPP & - & - & - & 44.0 & \textcolor{black}{N} \\
            \textcolor{black}{UperNet~\cite{xiao2018unified}} & \textcolor{black}{DeiT-S~\cite{touvron2021training}} & \textcolor{black}{-} & \textcolor{black}{43.0} & \textcolor{black}{-} & \textcolor{black}{41.0} & \textcolor{black}{Y} \\
            \textcolor{black}{UperNet~\cite{xiao2018unified}} & \textcolor{black}{Vim-S~\cite{zhu2024vision}} & \textcolor{black}{-} & \textcolor{black}{46.0} & \textcolor{black}{-} & \textcolor{black}{44.9} & \textcolor{black}{Y} \\
            \textcolor{black}{AFFormer-B~\cite{dong2023head}} & \textcolor{black}{-} & \textcolor{black}{-} & \textcolor{black}{3.0} & \textcolor{black}{49.6} & \textcolor{black}{41.8} & \textcolor{black}{Y} \\
            \textcolor{black}{RTLinearFormer~\cite{gu2025rtlinearformer}} & \textcolor{black}{-} & \textcolor{black}{14.7} & \textcolor{black}{24.0} & \textcolor{black}{\textbf{107.3}} & \textcolor{black}{35.7} & \textcolor{black}{Y} \\
			\midrule
			\midrule
			\textcolor{black}{\textbf{HyCTAS-L}} & HyCTAS-L & 72.7 & 43.2 & 71.8 & \textbf{44.9} & \textcolor{black}{N} \\
			\bottomrule[1.5pt]
	\end{tabular}\label{tab:ade20k}
\end{table}

\textbf{ADE20K.} We further evaluate our HyCTAS-L on the ADE20K validation set with the original image resolution of $640\times640$. From Table~\ref{tab:ade20k}, we can see that our model is much smaller and faster than others. Specifically, our HyCTAS-L achieves 44.9\% mIoU with 71.8 FPS, which is 0.9\% better than Auto-DeepLab~\cite{liu2019auto}. Compared to those models with a ResNet-101~\cite{he2016deep} encoder, we also achieve leading performance while using less than half the FLOPs. For example, the FLOPs of our HyCTAS-L is 72.7G, which is less than half of that of DeepLabV3+~\cite{chen2018encoder}. For the MobileNetV2~\cite{sandler2018mobilenetv2} encoder, we have similar FLOPs, but our HyCTAS-L surpasses DeepLabV3+~\cite{chen2018encoder}, PSPNet~\cite{zhao2017pyramid}, and FCN~\cite{long2015fully} by more than 10\%. Furthermore, our HyCTAS-L surpasses the latest SegFormer~\cite{xie2021segformer} by 7.3\% while also running around 40\% faster.

\textcolor{black}{Our approach shows improved results over recent efficient transformer models including UperNet with DeiT-S~\cite{touvron2021training} and Vim-S~\cite{zhu2024vision} backbones, as well as AFFormer-B~\cite{dong2023head}. These comparisons highlight the effectiveness of our search strategy in discovering optimal hybrid architectures that balance accuracy and efficiency better than handcrafted transformer-based designs.}

\textbf{COCO.} We employ the recent bottom-up panoptic segmentation framework~\cite{cheng2020panoptic} as our baseline. To ensure fast running speed, we remove the heavy context module and decoder module in Panoptic-Deeplab, which are built based on Atrous Spatial Pyramid Pooling (ASPP)~\cite{chen2017deeplab}. We further transfer the discovered architecture, i.e., HyCTAS-L, from Cityscapes to train on COCO. To ensure the inference speed, we set the test size to $640\times640$. Table ~\ref{tab:coco_val} reveals that without sacrificing much accuracy, our HyCTAS-L achieves an FPS of 63.9. This high speed is over 90\% faster than the closest competitor in FPS~\cite{yang2019deeperlab}. Meanwhile, we surpass Deeperlab-MNV2~\cite{yang2019deeperlab} by a large margin, \textit{i.e.}, 7.5\% PQ. Compared to Panoptic-Deeplab-MNV3~\cite{cheng2020panoptic}, our HyCTAS-L surpasses it on multiple performance points with 5.9\% PQ. When using multi-scale and flipping, we achieve the best PQ at $640\times640$ resolution, which is 1.6\% better than Panoptic-Deeplab-ResNet-50 and 3.0\% better than Deeperlab-Xception-71 ~\cite{yang2019deeperlab}.

\textcolor{black}{Our HyCTAS-L demonstrates competitive advantages over recent methods such as AFFormer-B~\cite{dong2023head} and RTLinearFormer~\cite{gu2025rtlinearformer} on the COCO dataset, demonstrating the generalization capability of our searched architectures across different benchmarks. These results further validate that our search-based approach can discover more effective hybrid designs than handcrafted transformer-based models.}

\begin{table}[!t]
	\vspace{-0.cm}
	\centering
	\caption{\textcolor{black}{Comparison of panoptic segmentation on the COCO \textit{val2017} set. $\dagger$ denotes speed on Tesla V100-SXM2 GPU; $\ddagger$ denotes multi-scale + flip evaluation. Entries without $\dagger$ are measured by us on RTX 3090 at 640$\times$640, batch size 1. For $\ddagger$ entries, a single fixed input size and FPS are not applicable; we denote Input Size as "MS+Flip (var)" and FPS as "n/a" for clarity. “-” denotes values not reported by the original papers. \textbf{Pretrain} indicates ImageNet pretraining status (Y/N/-).}}
	\small
	\begin{tabular}{c|c|c|c|c|c|c}
			\toprule[1.5pt]
			\multirow{2}{*}{Methods} & \multirow{2}{*}{Input Size} & Speed  & PQ & $\text{PQ}^{\text{Th}}$ & $\text{PQ}^{\text{St}}$ & \multirow{2}{*}{ \textcolor{black}{Pretrain}} \\
			& & {(FPS)}  & (\%) & (\%) & (\%) &  \\
			\midrule[1.5pt]
			$\text{Deeperlab}_{\text{MNV2}}$~\cite{yang2019deeperlab}& 641x641 & $17.2^{\dagger}$ & 27.9 & - & - & \textcolor{black}{Y} \\
			$\text{Deeperlab}_{\text{Xception-71}}$~\cite{yang2019deeperlab}& 641x641 & $10.6^{\dagger}$ & 33.8 & - & - & \textcolor{black}{Y} \\
			$\text{Panoptic-DeepLab}_{\text{MNV3}}$~\cite{cheng2020panoptic} & 641x641 & $13.5^{\dagger}$ & 29.5 & - & - & \textcolor{black}{Y} \\
			$\text{Panoptic-DeepLab}_{\text{ResNet-50}}$~\cite{cheng2020panoptic} & 641x641 & $6.0^{\dagger}$ & 35.2 & - & - & \textcolor{black}{Y} \\
            \textcolor{black}{AFFormer\text{-}B~\cite{dong2023head}} & \textcolor{black}{640x640} & \textcolor{black}{46.5} & \textcolor{black}{35.1} & \textcolor{black}{-} & \textcolor{black}{-} & \textcolor{black}{Y} \\
            \textcolor{black}{RTLinearFormer~\cite{gu2025rtlinearformer}} & \textcolor{black}{640x640} & \textcolor{black}{\textbf{150.2}} & \textcolor{black}{31.1} & \textcolor{black}{-} & \textcolor{black}{-} & \textcolor{black}{Y} \\
			\midrule
			\midrule
			\textbf{HyCTAS-L} & 640x640 & \textcolor{black}{63.9} & \textbf{35.4} & 36.8 & 33.1 & \textcolor{black}{N} \\
			\textbf{$\text{HyCTAS-L}^{\ddagger}$} & \textcolor{black}{MS+Flip (var)} & \textcolor{black}{n/a} & \textbf{36.8} & 38.5 & 34.5 & \textcolor{black}{N} \\
			\bottomrule[1.5pt]
	\end{tabular}
	\label{tab:coco_val}
\end{table}

\begin{table}[th]
	\centering
	\caption{Ablation study on the randomly sampled, locally searched, and NSGA-II searched architectures on the Cityscapes validation set.}
	\label{tab:random}
			\begin{tabular}{lcccc}
				\toprule[1.5pt]
				~ \multirow{2}{*}{Architectures} ~ & {FLOPs}  & \multicolumn{1}{c}{Params} & ~ mIoU ~\\
				& {(G)}  & \multicolumn{1}{c}{(M)} & (\%)
				\\ \midrule[1.5pt]
				Random-1 &$14.6\pm1.2$ & $1.5\pm0.2$ & $68.0\pm3.9$ \\
				\textcolor{black}{Local-1} & \textcolor{black}{11.7} & \textcolor{black}{1.6} & \textcolor{black}{70.5} \\
				\textbf{Searched-1} & 9.6 & 1.5 & 72.3  \\
				\midrule
				Random-2 & $31.6\pm1.0$ & $2.0\pm0.7$ & $73.6\pm1.8$ \\
				\textcolor{black}{Local-2} & \textcolor{black}{26.1} & \textcolor{black}{2.7} & \textcolor{black}{75.2} \\
				\textbf{Searched-2} & 22.4 & 3.2 & 77.1 \\
				\midrule
				Random-3 & $391.5\pm11.4$ & $32.0\pm4.8$ & $77.6\pm0.9$\\
				\textcolor{black}{Local-3} & \textcolor{black}{370.0} & \textcolor{black}{40.1} & \textcolor{black}{79.2} \\
				\textbf{Searched-3} & 335.1 & 43.2 & 80.9 \\
				\bottomrule[1.5pt]
		\end{tabular}
\end{table}

\begin{table}[th]
		\caption{Ablation study on the searched architectures on the Cityscapes validation set.}
		\label{tab:shift}
		\centering
			\begin{tabular}{lcccc }
				\toprule[1.5pt]
				~ \multirow{2}{*}{Architectures} ~ & {FLOPs}  & \multicolumn{1}{c}{Params}  & ~ mIoU ~\\
				& {(G)}  & \multicolumn{1}{c}{(M)}  & (\%)
				\\ \midrule[1.5pt]
				\textbf{Searched} & 29.4  & 2.4   & 77.1 \\
				\midrule
				+ SA $->$ Conv. & 23.1 & 2.8  & 75.6\\
				+ Conv. $->$ SA & 20.5 & 2.5  & 74.3\\
				+ Shift SA Position  & 23.1 & 2.8  & 75.2\\		
				+ Shift SA Branch  & 22.5 & 2.6 & 75.6 \\			
				\bottomrule[1.5pt]
		\end{tabular}
\end{table}

\subsection{Ablation}
\textcolor{black}{\textbf{Compared to Random and Local Search.} We compare HyCTAS to (i) random sampling and (ii) a multi-objective local search. For random sampling, we draw \textbf{5} seeds across 1–3 branches and retrain them. For local search, starting from the same \textbf{5} seeds, we run \textbf{32} iterations per seed; at each iteration we generate \textbf{5} neighbors by a single edit (operator swap or edge toggle). Selection is Pareto-aware: a neighbor replaces the incumbent only if it strictly dominates it in (mIoU\(\uparrow\), FLOPs\(\downarrow\)) for this FLOPs–mIoU analysis; when no neighbor dominates, we keep the incumbent and record the current non-dominated set. After search, we \emph{pool all evaluated Local Search candidates across all seeds and iterations} and compute the \emph{approximate Pareto front} (non-dominated set) with respect to (mIoU, FLOPs); we then re-train the resulting non-dominated models. Results for Random, Local Search, and HyCTAS are summarized in Table~\ref{tab:random}}

\textcolor{black}{\textbf{Selection of Search-1/2/3.}
From the \emph{approximate Pareto fronts} for each branch configuration (1-, 2-, and 3-path), we manually select Search-1/2/3 by jointly considering FLOPs, parameter count, and latency; these selections correspond to HyCTAS-S/M/L. More broadly, our method yields a family of models along these fronts, allowing practitioners to choose configurations that match their deployment priorities (e.g., maximizing mIoU for accuracy or minimizing latency for speed).
}

\textcolor{black}{\textbf{Selection of Local-1/2/3.} We pool all evaluated Local Search candidates across seeds and iterations and compute the \emph{approximate Pareto front} w.r.t.\ (mIoU, FLOPs). Local-1/2/3 are then chosen from this pooled front as the models whose FLOPs are \emph{closest} to the already selected Searched-1/2/3.}

\textbf{Robustness of Searching.} We further conducted an ablation study by intentionally changing components to verify the searching robustness of our HyCTAS-M model. As shown in Table \ref{tab:shift}, we distorted the searched architectures by replacing all self-attention with convolution, replacing all convolution with self-attention, shifting self-attention to another position, and shifting self-attention to another branch, i.e., moving it to a random position on the same stride. It can be seen that our searched models consistently perform better than these variants, proving their robustness.

In summary, the architecture searched by HyCTAS can find a great trade-off between performance, model size, and latency.


\begin{table}[th]
\centering
\caption{Experiments with different settings on the Cityscapes validation set. `SDP' denotes Scheduled Drop Path.}
    \begin{tabular}{lccccc}
        \toprule[1.5pt]
        \#  & Iter-186k & Iter-372k & SDP-0.1 & SDP-0.2 &  mIoU(\%) \\
        \midrule[1.5pt]
        1 & \cmark & & & & 71.5 \\
        2 & \cmark & & \cmark &  & 73.0 \\
        3 & \cmark & &  & \cmark & 73.3 \\
        4 &  & \cmark &  & \cmark& \textbf{73.5} \\
        \bottomrule[1.5pt]
    \end{tabular}
    \label{tab:hyperp}
\end{table}

\textbf{Retrain Settings.} From Table~\ref{tab:hyperp}, we can see that adding the scheduled drop path~\cite{liu2019auto} improves the mIoU by 1.5\%. When the scheduled drop path probability is set to 0.2, the performance is improved by 0.3\%. In \#4, we also study the effects of larger training iterations. Increasing the training iterations from 186K to 372K further improves the performance by 0.2\%. All reported results are obtained without ImageNet~\cite{deng2009imagenet} pretraining.

\textbf{Correlation Analysis.} We sampled 30 subnetworks, training each from scratch for 40,000 iterations. \textcolor{black}{We observe a positive rank and linear correlation between mIoU measured within the supernetwork and after retraining (Kendall's $\tau{=}0.51$, Pearson's $\rho{=}0.70$), suggesting that supernet evaluation is a useful proxy for ranking.} 

\begin{figure*}[h]
	\centering
	\includegraphics[width=1.\linewidth]{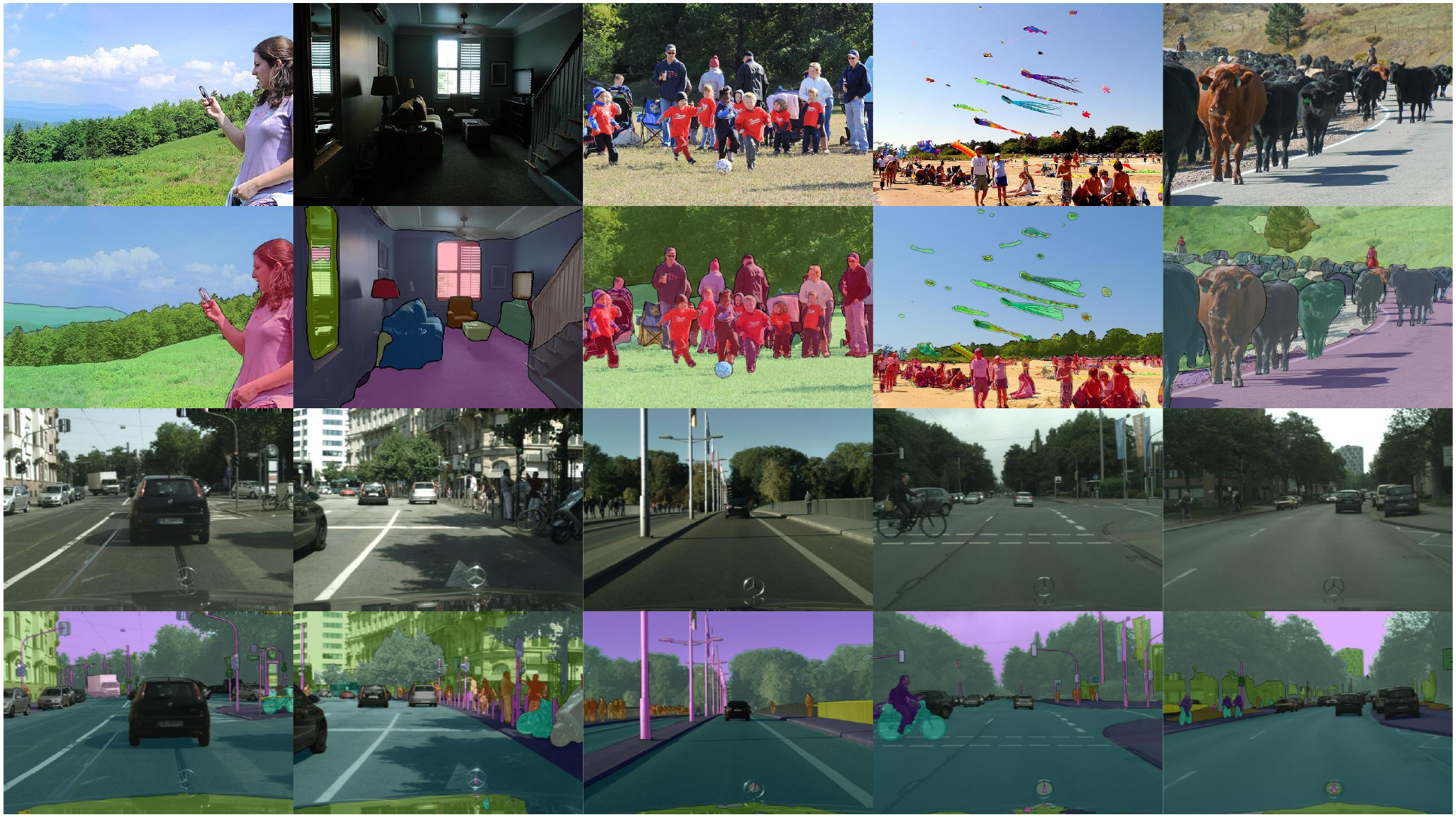}
	\caption{Visualization on COCO and Cityscapes validation set. The first row corresponds to COCO original images and the second row presents their semantic segmentation results. The third row corresponds to Cityscapes original images and the fourth row presents their semantic segmentation results. Best viewed in color. \textcolor{black}{Zoom in for details; predictions retain fine boundaries due to high-resolution branches.}}
	\label{fig:vis}
\end{figure*}

\subsection{Visualization}
\textcolor{black}{We visualize representative predictions in Figure \ref{fig:vis} on COCO and Cityscapes. Thanks to the high-resolution branches, HyCTAS retains thin structures and boundaries (e.g., poles, traffic signs) while memory-efficient attention supplies global context to disambiguate large ``stuff'' regions.
On COCO, HyCTAS-L yields crisp masks for overlapping instances and improves parsing of amorphous categories (road, sky) without sacrificing speed at 640$\times$640. On Cityscapes (1024$\times$2048), the method delineates small objects (pedestrians, riders) with a low rate of false merges.}

\section{Limitations}
Although our HyCTAS framework achieves state-of-the-art speed and accuracy trade-offs in semantic and panoptic segmentation, several limitations remain:

\begin{itemize}
    \item \textbf{Sensitivity to Noise and Occlusions.} Our method, like most deep architectures, may degrade under severe noise or occlusion. Future work could integrate robust loss functions or data augmentation targeted at severe occlusions.
    \item \textbf{Extreme Resolution Shifts.} While our multi-resolution branches handle moderate scale variations, extremely large input images or drastic changes in scale may still pose memory challenges, even with our memory-efficient attention.
    \item \textbf{Domain Generalization.} We have mainly evaluated on well-known datasets such as Cityscapes and ADE20K. In real-world applications (e.g., certain industrial or medical imaging domains), domain shifts could undermine performance if the training distribution differs substantially.
    \item \textcolor{black}{\textbf{Search cost and hardware dependency.} Although the search is relatively efficient, the supernet pretraining plus NSGA-II still requires notable GPU hours, and the latency objective is calibrated to a specific GPU/driver stack. Architectures on the \emph{approximate Pareto front} may shift if the deployment hardware changes (e.g., mobile DSP vs. server GPU).}
    \item \textcolor{black}{\textbf{Calibration and small-batch effects.} Latency and accuracy during search are measured with batch size 1 and BatchNorm re-estimation on a subset; distribution shifts at deployment (e.g., different preprocessing or batch size) can slightly change the accuracy–latency trade-off.}
    \item \textcolor{black}{\textbf{Transferability across tasks.} While we show promising panoptic results, transfer to tasks with different optimization targets (e.g., detection or pose) is not guaranteed; the search objectives would likely need retuning.}
\end{itemize}

\textcolor{black}{To mitigate these issues, future work will (i) incorporate portable latency proxies (e.g., per-operator cost models) for cross-hardware robustness; (ii) explore differentiable or predictor-guided multi-objective search to further reduce GPU hours; and (iii) extend the search space with domain-adaptive modules and normalization layers that are less sensitive to batch effects.}

We plan to investigate these aspects in future research, potentially by extending our search space to incorporate more robust modules or domain-adaptive strategies.

\section{Conclusion}
In this paper, we propose a novel HyCTAS framework to search for efficient networks with high-resolution representation and attention. We introduce a lightweight convolution module to reduce computation costs while preserving high-resolution information and a memory-efficient self-attention module to capture long-range dependencies. By combining a multi-branch search space and an \emph{evolutionary computation-based approach (NSGA-II)} with multiple objectives, our searched architectures not only fuse features from selective branches for efficiency but also maintain high-resolution representations for performance. Additionally, the discovered architecture provides insights into how to use self-attention when designing new networks. Extensive experiments demonstrate that HyCTAS outperforms previous methods in both semantic segmentation and panoptic segmentation tasks while significantly improving inference speed. Future work will focus on applying this approach to more challenging tasks, such as key-point localization and object detection.

\textcolor{black}{In summary, HyCTAS conducts a single multi-objective search to produce an \emph{approximate Pareto set} of hybrid convolution–attention architectures that operate at high resolution and deliver strong accuracy–latency trade-offs on Cityscapes, ADE20K, and COCO, all trained from scratch. The search further reveals systematic placement patterns of attention across branches (shallower for single-branch, deeper for multi-branch) that can inform future network design.}

\textcolor{black}{The primary contribution is the \emph{NAS strategy and search space} for multi-branch high-resolution segmentation with explicit latency as an objective; the lightweight convolution and memory-efficient attention are pragmatic operator choices that enable feasible search rather than architectural novelties.}

\textcolor{black}{Regarding limitations and outlook, the latency estimates are hardware-specific and the search still requires GPU hours. Future directions include portable latency models and predictor-guided search to reduce cost, analyzing attention placement under different objective weights, and extending HyCTAS to additional tasks.}

\section*{Author Contributions}
\noindent
\textbf{H.~Yu} conceived the initial idea, designed the supernet, and conducted the majority of experiments.
\textbf{C.~Wan} and \textbf{X.~Dai} contributed to the multi-objective search algorithm implementation and code maintenance.
\textbf{M.~Liu}, \textbf{D.~Chen}, and \textbf{B.~Xiao} provided guidance on memory-efficient self-attention design and revised the manuscript.
\textbf{Y.~Huang}, \textbf{Y.~Lu}, and \textbf{L.~Wang} brought expertise in high-resolution networks, coordinated the experimental setup, and offered theoretical insights on multi-branch architectures.
All authors reviewed the paper, discussed experiments, and approved the final manuscript. All authors agree to the author list and order and are accountable for all aspects of the work.

\bibliographystyle{elsarticle-num} 
\bibliography{main}
\end{document}